\documentclass[10pt,twocolumn,letterpaper]{article}

\usepackage{wacv}              
\usepackage[accsupp]{axessibility}  

\usepackage{graphicx}
\usepackage{amsmath}
\usepackage{amssymb}
\usepackage{booktabs}
\usepackage[normalem]{ulem}
\usepackage{url}
\useunder{\uline}{\ul}{}
\usepackage{enumitem}

\usepackage[pagebackref,breaklinks,colorlinks]{hyperref}

\usepackage[capitalize]{cleveref}
\crefname{section}{Sec.}{Secs.}
\Crefname{section}{Section}{Sections}
\Crefname{table}{Table}{Tables}
\crefname{table}{Tab.}{Tabs.}

\usepackage{tabularx}

\def\wacvPaperID{1438} 
\def\confName{WACV}
\def\confYear{2024}

\begin{document}

\title{EvDNeRF: Reconstructing Event Data with Dynamic Neural Radiance Fields}

\author{Anish Bhattacharya\\
University of Pennsylvania\\
Philadelphia, PA 19104
\\
{\tt\small anish1@seas.upenn.edu}
\and
Ratnesh Madaan\\
Microsoft\\
Redmond, WA 98052\\
{\tt\small ratneshmadaan@gmail.com}
\and
Fernando Cladera\\
University of Pennsylvania\\
Philadelphia, PA 19104\\
{\tt\small fclad@seas.upenn.edu}
\and
Sai Vemprala\\
Scaled Foundations\\
Redmond, WA 98052\\
{\tt\small saihvemprala@gmail.com}
\and
Rogerio Bonatti\\
Microsoft\\
Redmond, WA 98052\\
{\tt\small rbonatti@microsoft.com}
\and
Kostas Daniilidis\\
University of Pennsylvania\\
Philadelphia, PA 19104\\
{\tt\small kostas@cis.upenn.edu}
\and
Ashish Kapoor\\
Scaled Foundations\\
Redmond, WA 98052\\
{\tt\small ashish@scaledfoundations.ai}
\and
Vijay Kumar\\
University of Pennsylvania\\
Philadelphia, PA 19104\\
{\tt\small kumar@seas.upenn.edu}
\and
Nikolai Matni\\
University of Pennsylvania\\
Philadelphia, PA 19104\\
{\tt\small nmatni@seas.upenn.edu}
\and
Jayesh K. Gupta\\
Poly Corporation\\
Bellevue, WA 98004\\
{\tt\small jayesh@withpoly.com}
}
\maketitle


\begin{abstract}
\vspace{-.4cm}
   We present EvDNeRF, a pipeline for generating event data and training an event-based dynamic NeRF, for the purpose of faithfully reconstructing eventstreams on scenes with rigid and non-rigid deformations that may be too fast to capture with a standard camera. Event cameras register asynchronous per-pixel brightness changes at MHz rates with high dynamic range, making them ideal for observing fast motion with almost no motion blur. Neural radiance fields (NeRFs) offer visual-quality geometric-based learnable rendering, but prior work with events has only considered reconstruction of static scenes. Our EvDNeRF can predict eventstreams of dynamic scenes from a static or moving viewpoint between any desired timestamps, thereby allowing it to be used as an event-based simulator for a given scene. We show that by training on varied batch sizes of events, we can improve test-time predictions of events at fine time resolutions, outperforming baselines that pair standard dynamic NeRFs with event generators. We release our simulated and real datasets, as well as code for multi-view event-based data generation and the training and evaluation of EvDNeRF models \footnote{\url{https://github.com/anish-bhattacharya/EvDNeRF}}.
\end{abstract}


\setlength{\belowcaptionskip}{-10pt}

\begin{figure*}[t]
    \centering
    \includegraphics[width=0.8\linewidth]{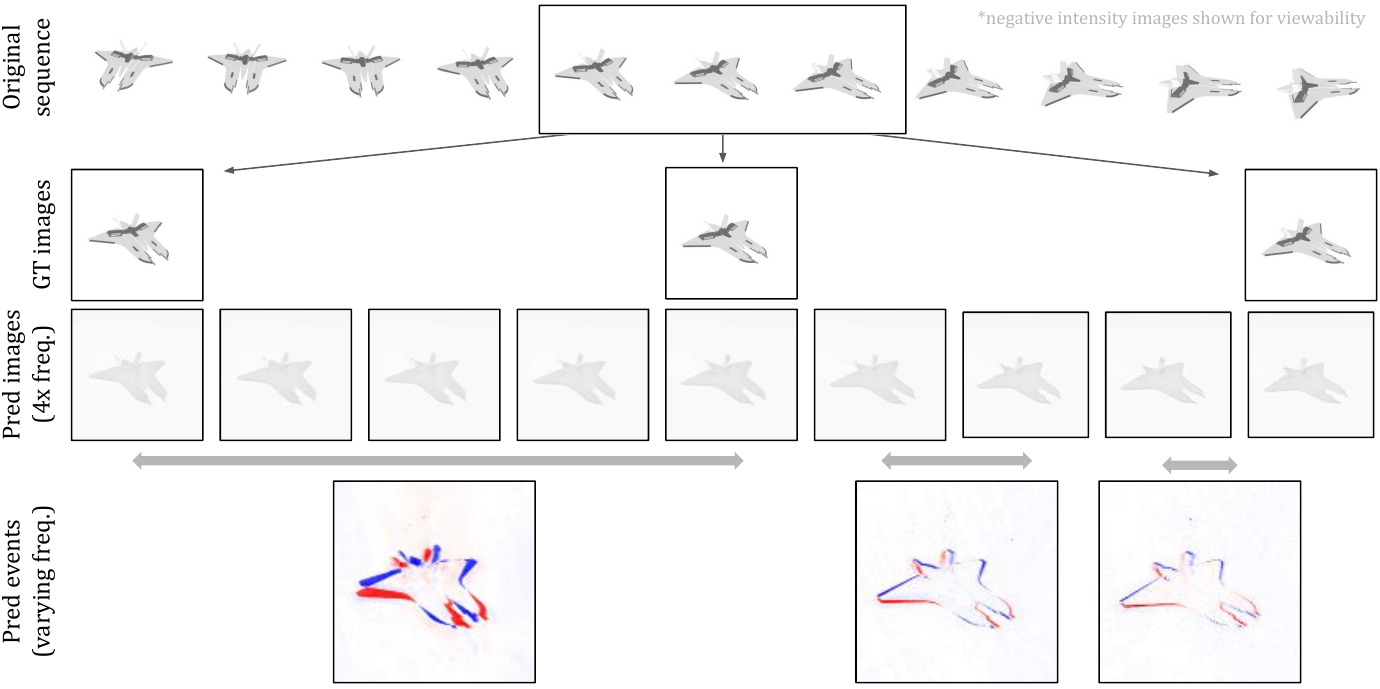}
    \caption{Evaluation of EvDNeRF on Jet-Down dataset during unseen camera motion around object. Ground truth images are sparse, but we can query the network for intensity images at a higher rate. We also vary the temporal resolution of predicted events, showing the original, 4x, and 8x resolution in the final row.}
    \label{fig:pred-ims-evims}
\end{figure*}


\vspace{-.3cm}

\section{Introduction}
\label{sec:intro}


Simulation of 3D scenes based on discrete observations of 2D images is a challenging problem in computer graphics and robotics. Occlusions, shadows, imperfect state estimation, and perspective distortion make this a difficult task for classical graphics modeling methods. Recently, neural radiance fields (NeRF)~\cite{mildenhall2020nerf} have emerged as a promising solution to these challenges.
This method can take in a set of images from known camera viewpoints, and generate views of the reconstructed scene from novel viewpoints with high visual fidelity. This work has been extended to dynamic scenes with complex textures and structures~\cite{pumarola2020dnerf, li2021nsff, tretschk2021nrnerf, du2021neural}, which can be useful for planning and simulation. However, these methods work under the assumption of little to no motion blur and favorable lighting conditions, and fail in many dynamic scenes that break such constraints, as is often the case when generating 3D reconstructions of scenes in the wild. For example, it would be extremely difficult to create usable training images from a camera carried by a jogging human or a flying quadrotor, or of scenes with high-speed motion such as a projectile. This has motivated recent work training NeRF models from event cameras~\cite{rudnev2023eventnerf, klenk2023enerf, hwang2023evnerf}.

Event-based cameras, or neuromorphic cameras, are small and lightweight vision sensors containing pixels which fire independently at MHz frequencies. Each pixel detects changes in brightness, triggering a positive or negative ``event" at that pixel location depending on a tunable event threshold. This asynchronous datastream of events (interchangeably events, event data, eventstream) is in stark contrast to the synchronous frame-based data from a 30-60Hz CMOS camera. While challenging to use for classical computer vision tasks, event cameras have now been used in the computer vision and robotics community to reconstruct images and video~\cite{munda2018real, scheerlinck2018continuous, scheerlinck20wacv, rebecq2019e2vid, pan2019bringing, pan2020high, rebecq2019high}, perform state estimation~\cite{mueggler2018continuous, chen2023esvio, dimitrova2020towards, zuo2022devo, zhou2021event, rebecq2016evo, kueng2016low, hadviger2021feature, kim2016real}, control robots~\cite{falanga2020dynamic, dimitrova2020towards, eguiluz2021fly, sanket2020evdodgenet, kaiser2016towards, wang2022ev}, and more \cite{gallego2020event}. These works use a variety of methods, from filtering to learning-based. Event cameras also have a high dynamic range, reaching 120dB, more than double that of a standard camera (event frequencies drop to 10-100kHz in low light). The detection of brightness changes naturally lends itself to motion detection and modeling without the challenge of parsing out static backgrounds.

Generating simulated events from a dynamic scene has been done by processing RGB video with either learned methods or an accurate model of an event camera sensor~\cite{gehrig2020vid2e, delbruck2020v2e, garcia2016pydvs}. However, these methods cannot generalize to novel viewpoints or temporal/spatial resolutions. Models trained on NeRF backbones, however, can provide a strong geometric prior to make such generalizations possible. Initial work training NeRFs from events~\cite{rudnev2023eventnerf, hwang2023evnerf, klenk2023enerf} considers static scenes, with single trajectory camera motion creating the eventstream. In contrast to these works, we train dynamic NeRFs on event data of moving scenes. We present results on both simulated and real data, with single objects exhibiting highly dynamic trajectories as well as non-rigid deformations, and scenes with multiple objects and occlusions. To the best of our knowledge, we are the first to train dynamic NeRF models on event data, which advances the state-of-the-art in generative events simulators, thereby enabling future event-based applications. \textbf{Our contributions are as follows:}

\begin{itemize}
    \itemsep0em
    \item EvDNeRF, the first dynamic NeRF trained from event data; we show generalization to event predictions at unseen, fine time resolutions and camera motion.
    \item A data generation procedure of multi-view eventstreams from rigid and non-rigid dynamic scenes in simulation and real-world.
    \item Demonstration and analysis of transfer learning an EvDNeRF to an unseen object.
    \item Open-source code and datasets to reproduce our results or train new models.
\end{itemize}


\section{Related works}
\label{sec:related-works}


\noindent \textbf{Modeling dynamic scenes.} Modeling moving scenes from vision data alone is a challenging open task in computer vision. Since the problem is constrained only by 2D images collected over time, the task becomes more constrained as the amount of diverse data increases. The ability to train models in a self-supervised fashion with images was enabled by neural implicit rendering, starting with \cite{mildenhall2020nerf}. Extensions to dynamic scenes with rigid and non-rigid deformations largely have fallen into two categories, one training modular deformation or flow networks \cite{pumarola2020dnerf, tretschk2021nrnerf, du2021neural} and another directly training higher-dimensional architectures \cite{li2021nsff, ost2021neural, park2021nerfies, park2021hypernerf} with more complex optimization landscapes.


\noindent \textbf{Learning static NeRFs from event data.} There has recently been some work on training traditional, static NeRF models from event camera data \cite{rudnev2023eventnerf, hwang2023evnerf, klenk2023enerf}. Rudnev, \etal (2023) does so with color event cameras to reconstruct objects in color and depth. Since the scene is static, the event camera is rotated around the object to generate the motion necessary for events. A combination of \textit{positive} sampling of non-zero event pixels and \textit{negative} sampling of background pixels makes for efficient NeRF training.
Hwang, \etal (2023) uses a deadzone event reconstruction loss that has zero gradient between target event threshold discretizations. The positive and negative thresholds themselves are jointly learned, while regularizing distance from \textit{a priori} known approximate values. They train on evenly-sized time windows of events and add small amounts of random noise during training to generate a model that is robust to event camera noise, which can occur in low-light conditions.
Klenk, \etal (2023) similarly trains a NeRF from eventstreams, but employs additional loss terms (a) regularizing predictions in areas of extended-durations of no events, and (b) comparing predicted intensity images to ground truth blurred RGB frames of the scene to reconstruct colored images.
These works attempt to generate high visual quality images of a static scene by training NeRFs on eventstreams and using an affine transformation on the resulting image predictions (or directly optimizing on blurred image frames) to correct the image brightness. In contrast, our work focuses on making accurate predictions of eventstreams themselves, on dynamic scenes with rigid and non-rigid deformations, with test-time generalization to queries of novel event batch size and viewing angle or camera motion.

\noindent \textbf{Simulating event-based data.} A small number of events simulators have been released that approach the problem of modeling this unique data in different ways. Some of these attempt to model the unique characteristics of the neuromorphic camera and its parameters to a high accuracy \cite{delbruck2020v2e, joubert2021event}. This may help reduce the sim-to-real gap when transferring algorithms developed in simulation to real-world. Another work \cite{mueggler2017event} utilizes a surface of active events \cite{benosman2013event} to calculate events from the high-fidelity Blender renderer. ESIM \cite{rebecq2018esim} (part of the Vid2E pipeline \cite{gehrig2020vid2e}) benefits from utilizing a deep-learning-based video upsampler \cite{jiang2018super} to generate continuous streams of events from sparse images. However, without proper upsampling this method struggles to simulate events at arbitrary intermediate timesteps, and still may contain patchy artifacts. Unlike our work, none of these simulators leverage geometric understanding of the scene, and therefore cannot simulate events from viewpoints outside the provided data.


\section{Methods}
\label{sec:methods}

We aim to build a dynamic neural radiance field from an eventstream, for predicting the expected events of a dynamic, bounded scene. To serve as an events simulator, we would like this model to be able to predict events on fine temporal resolutions and novel viewpoints outside the training data, from a variety of datasets with occlusions, shadows, scene dynamics, and multiple objects. The methods presented below leverage the asynchronous, discrete, very fast nature of event cameras to accomplish these goals better than traditional image-based simulators.

\subsection{Problem formulation}
\label{sec:problem-formulation}

\setlength{\belowdisplayskip}{2pt} \setlength{\belowdisplayshortskip}{0pt}
\setlength{\abovedisplayskip}{2pt} \setlength{\abovedisplayshortskip}{0pt}

A single event is represented as $e_k = (t_k, x_k, y_k, p_k)$, denoting a brightness change registered by an event camera at time $t_k$, pixel location $(x_k, y_k)$ in the camera frame, with polarity $p_k \in \{-1, +1\}$. The polarity of an event indicates a positive or negative change in logarithmic illumination, quantized by positive and negative thresholds $C^{\pm}$. The change in brightness between two timesteps can be calculated from intensity images $B_t$.
\begin{align}
    \Delta L_{(t_k, x_k, y_k)} &= L(t_k, x_k, y_k) - L(t_{k-1}, x_k, y_k) \label{eq:delta-L} \\
    e_p &= \left\{
        \begin{array}{ll}
            -1 ,& \quad \text{if } \Delta L \leq C^- \\
            +1 ,& \quad \text{if } \Delta L \geq C^+
        \end{array}
    \right.
    \label{eq:ev-p} \\
    \text{where} \quad L &= \log (B). \label{eq:log-B}
\end{align}
Our pipeline approaches $\Delta \hat{L}_{t_k}$ estimation via Equation \ref{eq:delta-L}, by generating intensity image predictions $\hat{B_t}$ at two times $t_k$ and $t_{k-1}$ and using Equation \ref{eq:log-B}. We can then determine the number $\hat{N}_{evs}$ of predicted events at a pixel location during that time window:
\begin{align}
    \hat{N}_{evs,(x_k,y_k)} &= \left\{
        \begin{array}{ll}
            \left \lfloor \frac{\Delta \hat{L}_{(x_k,y_k)}}{C^+} \right \rfloor ,& \; \text{if } \Delta L \geq 0 \vspace{0.2cm}  \\
            \left \lfloor \frac{\Delta \hat{L}_{(x_k,y_k)}}{|C^-|} \right \rfloor ,& \; \text{if } \Delta L < 0.
        \end{array}
    \right.
    \label{eq:counting-events}
\end{align}
Given a supervisory eventstream $E_N$ and Equation \ref{eq:ev-p}, we can calculate each pixel's true logarithm brightness changes for a given time window $t_{k-1} < \tau < t_k$ batch of events (interchangeably event batch, sliced events, and so on) by a per-pixel summation:
\begin{align}
    \Delta L_{(t_k, x_k, y_k)} = \sum_{p  \in \{-1, +1\}} C^p E_{\{ t_{k-1} < \tau < t_k \} , x_k, y_k, p}
    \label{eq:events-to-delta-L}
\end{align}
Therefore, we can supervise EvDNeRF training between our predicted $\Delta \hat{L}_{t_k}$ and the true $\Delta L_{t_k}$ by Equation \ref{eq:events-to-delta-L} over all image pixels.

\begin{figure}
  \centering
  \begin{subfigure}{0.45\linewidth}
    \includegraphics[width=1.0\linewidth]{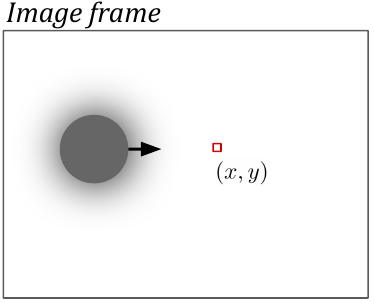}
    \caption{Illustration of an object with a smooth edge passing rightward over a pixel.}
    \label{fig:event-discretization-image}
  \end{subfigure}
  \hfill
  \begin{subfigure}{0.49\linewidth}
    \includegraphics[width=1.0\linewidth]{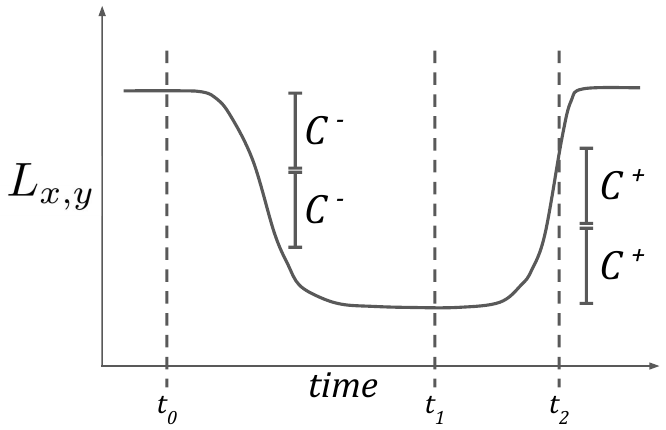}
    \caption{Logarithm brightness change at pixel $(x,y)$.}
    \label{fig:event-discretization-plot}
  \end{subfigure}
  \caption{Demonstration of how time window discretizations can influence the count of events between timestep pairs. The time window $(t_0, t_1)$ produces two negative events, whereas $(t_0, t_2)$ produces no events.}
  \label{fig:event-discretization}
\end{figure}

An important note is that the calculated $\hat{N}_{evs}$ is a minimum count; as the time window $t_k - t_{k-1}$ increases, the predicted events form a discretized, aliased representation of the underlying smooth brightness changes occurring in the scene. For demonstration, Figure \ref{fig:event-discretization} shows how measurements between timesteps $(t_0,t_1)$ would result in two negative events at the chosen pixel, but measurements between $(t_0,t_2)$ would result in no events. Given this observation, we purposely vary the time window of batched events during the training of EvDNeRF, and notably find that this improves our generation of finely-sliced events at test time. Note that while event cameras may not produce additional events at an event pixel during its refractory period, the scale of our time windows are generally an order of magnitude larger, so we disregard this parameter.


\subsection{Event data generation}
\label{sec:event-data-generation}

For both the simulation and real-world datasets we emulate a motion capture-like scenario of multiple event cameras capturing time synchronized datastreams from equally-spaced viewpoints. In simulation, we use an open-source simulator and renderer for data generation paired with an event generation package; in real-world we use one event camera and re-capture a repeatable scene motion at various viewing angles.


\begin{figure*}
  \centering
  \begin{subfigure}{0.48\linewidth}
    \includegraphics[width=1.0\linewidth]{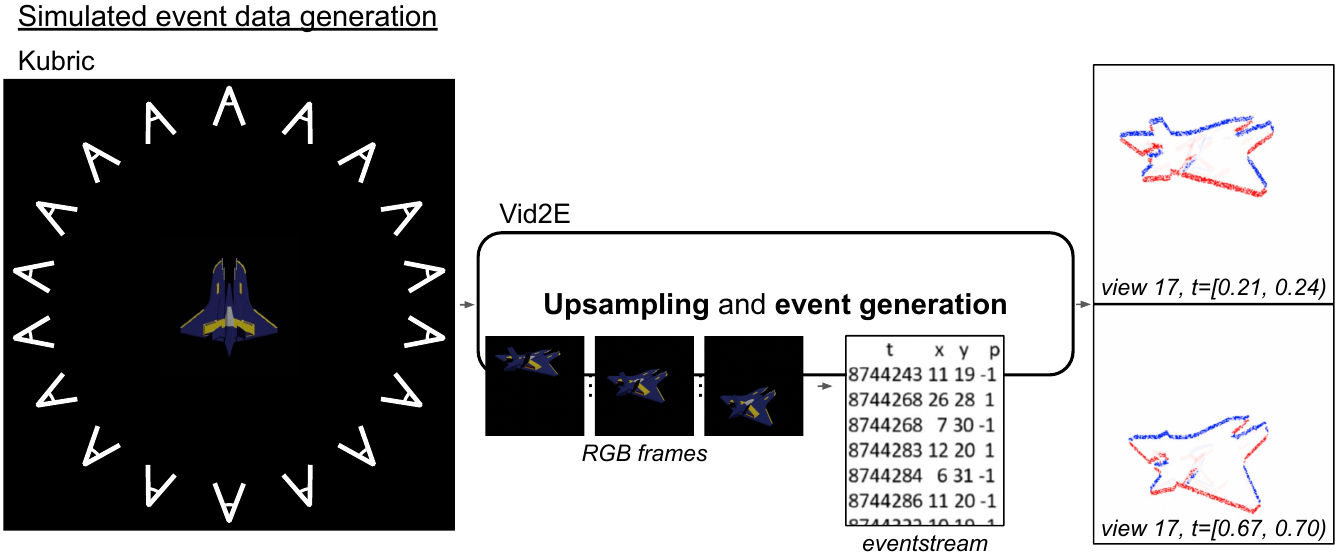}
    \caption{Generating simulated eventstreams from rendered images. We take blurred grayscale images and eventstreams as the data available to models for training.}
    \label{fig:datagen-sim}
  \end{subfigure}
  \hfill
  \begin{subfigure}{0.48\linewidth}
    \includegraphics[width=0.98\linewidth]{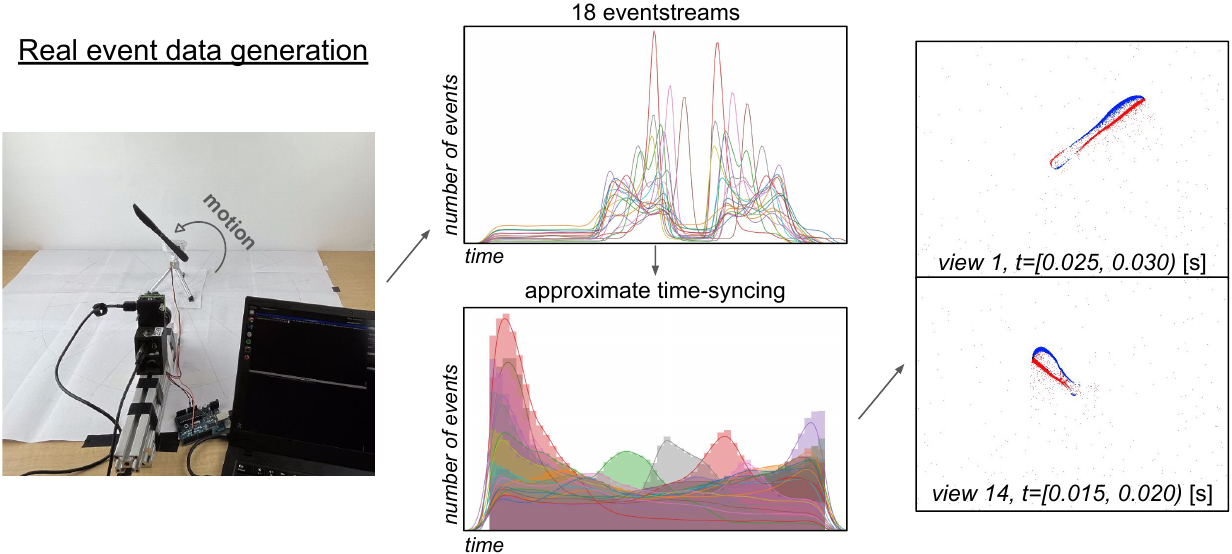}
    \caption{Real eventstreams from an event camera from multiple viewpoints. We take the eventstreams as the data available to models for training.}
    \label{fig:datagen-real}
  \end{subfigure}
  \vspace{0.1cm}
  \caption{Event data generation for simulated and real sequences. Note that in simulation, timestamps are unitless and range from [0, 1.0], whereas in real-world, timestamps are in seconds and range from [0, 0.5][\textit{s}].}
  \label{fig:datagen}
\end{figure*}

\noindent \textbf{Simulation.} We use the open source data generation package Kubric~\cite{greff2021kubric} to generate synthetic, photorealistic videos of dynamic objects using the renderer Blender. These scenes are bounded within a [4m, 4m, 4m] box with no background textures. The objects are pulled from the ShapeNet online dataset of textured meshes~\cite{chang2015shapenet} and each motion trajectory (\eg Down, Spiral, Land) is defined by setting the absolute pose of the object at discrete, sampled timesteps. 18 camera viewing angles (20$^\circ$ separation) around the scene capture time-synchronized images without blur or noise. These images are passed into the Vid2E package~\cite{gehrig2020vid2e} which upsamples each sequence via Super-SloMo~\cite{jiang2018super} and produces a continuous eventstream from each viewpoint using ESIM~\cite{rebecq2018esim}. Figure \ref{fig:datagen-sim} outlines this process. For quantitative validation, we generate eventstreams from 18 viewpoints offset from training viewpoints by 10$^\circ$.

\noindent \textbf{Real world.} Since we require a high number of viewpoints of a dynamic scene but do not have access to a large array of timesynced event cameras, we create a highly repeatable scene motion. An example is seen in Figure \ref{fig:datagen-real}, where a knife rotates on a servo from an initial state to the final position seen in the image. The tripod-mounted servo is placed at 18 orientations relative to the event camera and the servo is triggered via an Arduino microcontroller. Eventstream timestamps are plotted in a histogram, where a sudden peak in events indicates the start of a scene motion; this serves as the synchronization timestamp between all viewpoints. Once all viewpoints are time-synchronized, the eventstreams (each a N$\times$4 array) can be batched by arbitrary time windows for training EvDNeRF.

\subsection{Training dynamic neural radiance fields from events}

\noindent \textbf{Dynamic NeRF model.} We use DNeRF~\cite{pumarola2020dnerf} as the backbone for EvDNeRF. This architecture incorporates a deformation network $\Psi_t(\mathbf{x}, t)$ that predicts the displacement of any 3D point in space $\mathbf{x}$ at time $t$ relative to its position in some canonical scene, typically taken to be the scene reconstruction at $t=0$. Predicted displacements of each query point are used to feed deformation-compensated points into the canonical NeRF model $\Psi_x(\mathbf{x}, (\theta, \phi))$ which predicts color and density at these points, dependent on viewing angle $(\theta, \phi)$, to be fed into a backwards differentiable rendering function. We assume our dynamic scene is contained within a 3D origin-centered box of size $[b, b, b]$, and that we only deal with intensity (\ie grayscale) images since most event cameras are not color (thereby making $\mathbf{c}$ a scalar, though we denote it as a vector for generality).
\begin{align*}
    &\Psi_t(\mathbf{x}, t) = \Delta \mathbf{x} \text{ , where }\; \mathbf{x} \in \mathbb{R}^3 \; \text{ and } \; \frac{-b}{2} < \mathbf{x}_i < \frac{b}{2}.\\
    &\Psi_x\left(\mathbf{x}, (\theta, \phi)\right) = (\mathbf{c}, \sigma)
\end{align*}
The NeRF network $\Psi_x$~\cite{mildenhall2020nerf} learns volumetric density by simulating the geometry of a renderer and casting rays from multiple annotated viewpoints. Both $\Psi_t$ and $\Psi_x$, composed of fully-connected feedforward networks (\ie multilayer perceptrons, or MLPs), are able to learn continuous deformation and density models in 3D space, respectively, by being paired with differentiable rendering~\cite{max1995optical} and positional encoding of query points prior to input into the neural network~\cite{tancik2020fourier}.

The rendering function accumulates $S$ colors and densities at points $\mathbf{x}_S$ along a ray $\mathbf{r}$ to obtain a pixel's color $C(\mathbf{r})$:
\begin{align*}
    C(\mathbf{r}) &= \sum_{i}^S T_i \alpha_i \mathbf{c}_i\\
    \text{where } T_i &= \exp \left( - \sum_{j=1}^{i-1} \sigma_j \| \mathbf{x_{j+1}} - \mathbf{x_{j}} \| \right) \text{ ,}\\
    \alpha_i &= 1 - \exp(-\sigma_i \| \mathbf{x_{i+1}} - \mathbf{x_{i}} \|).
\end{align*}
The ray $\mathbf{r}$ is a unit vector with origin at camera origin and direction through a pixel center. The transmittance for a point along the ray, $T_i$, is determined by accumulated densities from the camera origin until that point. The alpha value $\alpha_i$ is calculated at each point only relative to the distance to a neighboring sample point.

Using this rendering function and the DNeRF backbone, as we described in Section \ref{sec:problem-formulation}, we predict intensity images at two time instances and calculate the continuous-valued log brightness change $\Delta \hat{L}_{t_k}$ (Equation \ref{eq:delta-L}), which can then be compared to the ground truth discrete-valued $\Delta L_{t_k}$ (Equation \ref{eq:events-to-delta-L}). During training, we sample 50\% of rays positively from event pixels and 50\% randomly.

\noindent \textbf{Varied batching of events.} We vary the batch size of events during training to enable EvDNeRF to make fine temporal predictions at test time, as motivated in Section \ref{sec:problem-formulation}. We start training with batches of events equal in time window to the framerate of the image stream (\eg 32 frames for simulated scenes), and halve the batch size multiple times during training, thereby repeatedly doubling the temporal resolution of our supervisory signal.

\noindent \textbf{Loss function.} An event reconstruction loss term $\mathcal{L}_{ev}$ is formulated as a piecewise loss function calculated at each pixel location to accommodate the discrete nature of threshold-based events.
\begin{align*}
    \mathcal{L}_{ev,\;xy} &= \left\{
        \begin{array}{ll}
            & 0 ,\; \text{if } \Delta L_{xy} \leq \Delta \hat{L}_{xy} < \Delta L_{xy} + C^{\pm} \vspace{0.2cm}  \\
            & \left\| \Delta \hat{L}_{xy} - \left( \Delta L_{xy} + \frac{C^{\pm}}{2} \right) \right\|_2^2 ,\; \text{otherwise. }
        \end{array}
    \right.
\end{align*}
The positive or negative threshold $C^{\pm}$ is determined by the polarity of the ground truth value $\Delta L_{xy}$. Within the $C^{\pm}$-width bin of any particular discretized, ground truth, log-brightness-change $\Delta L_{xy}$, the loss of prediction $\Delta \hat{L}_{xy}$ is valued $0$. However, outside this bin, the loss is valued as a squared distance to the center of that bin.



\section{Results}
\label{sec:results}

\begin{figure}
    \centering
    \includegraphics[width=1.0\linewidth]{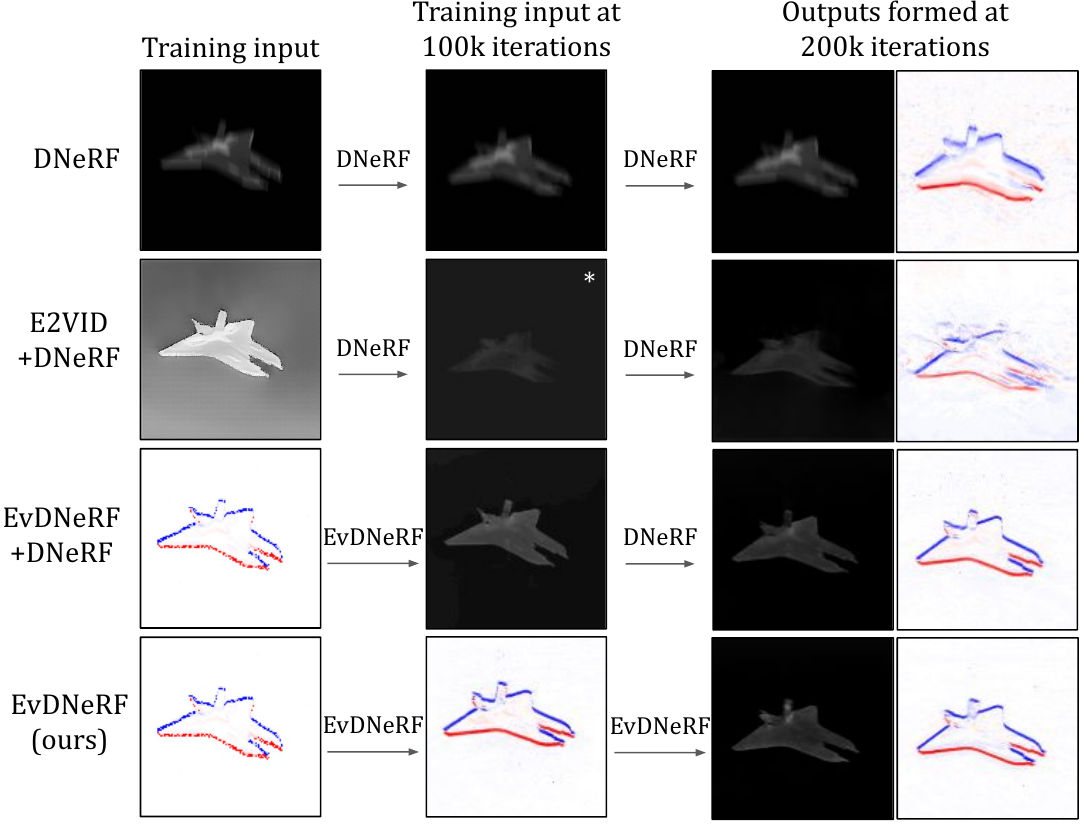}
    \caption{Baselines and EvDNeRF training explained, by visualizing sample data at the first training step, training midpoint, and end of training (output). $\xrightarrow{\text{DNeRF}}$ and $\xrightarrow{\text{EvDNeRF}}$ indicate 100k training steps of either method; EvDNeRF trains directly from event batches while DNeRF trains from images (baseline method \textit{DNeRF} trains from blurred images). \\(\textit{*} contrast improved for viewability.)}
    \label{fig:baselines-illustrated}
\end{figure}

We present events reconstruction images from EvDNeRF across various test-time queries. Each events reconstruction is a pseudoframe representing the collection of events predicted between two query timestamps, coalesced into an image where the intensity of pixel color represents the number of events at a pixel location (red indicates positive events, blue indicates negative events). Equations \ref{eq:delta-L} and \ref{eq:counting-events} describe the event counting process from network predictions. We show various events reconstructions for training samples, very fine untrained time slices, unseen viewpoints, and nonzero camera motion, with comparison to ground truth samples when available. While the problem of reconstructing events from a geometric model of a dynamic scene does not have other works with which to make direct comparison, we form baselines to be competitive with our model across certain metrics. We found EvDNeRF to be more unstable to train than DNeRF, likely due to the task of training from the finite-difference-like event data, and that it does not easily reconstruct visual-quality images or depth maps. Therefore, while the most simple baseline model is formed from training DNeRF on the available motion-blurred images (\textit{DNeRF}), we also attempt to utilize DNeRF to possibly ``smoothen'' out the results of EvDNeRF for improved events reconstruction; to this end, \textit{EvDNeRF+DNeRF} uses image outputs from EvDNeRF at 100k iterations of training for another 100k iterations of DNeRF training. We also use E2VID to train a DNeRF model from images reconstructed from the eventstream (\textit{E2VID+DNeRF}). These methods are illustrated in Figure \ref{fig:baselines-illustrated}, with examples of training input types, training midpoint data types, and reconstructed validation outputs. Finally, as we aim to reconstruct the eventstream itself, we also tested using VID2E on the output of \textit{E2VID+DNeRF} to reconstruct high-quality eventstreams; however, this returned high levels of background events, and therefore is not pictured here (details in the supplementary).

\subsection{Events reconstruction}

\begin{figure}
    \centering
    \includegraphics[width=0.7\linewidth]{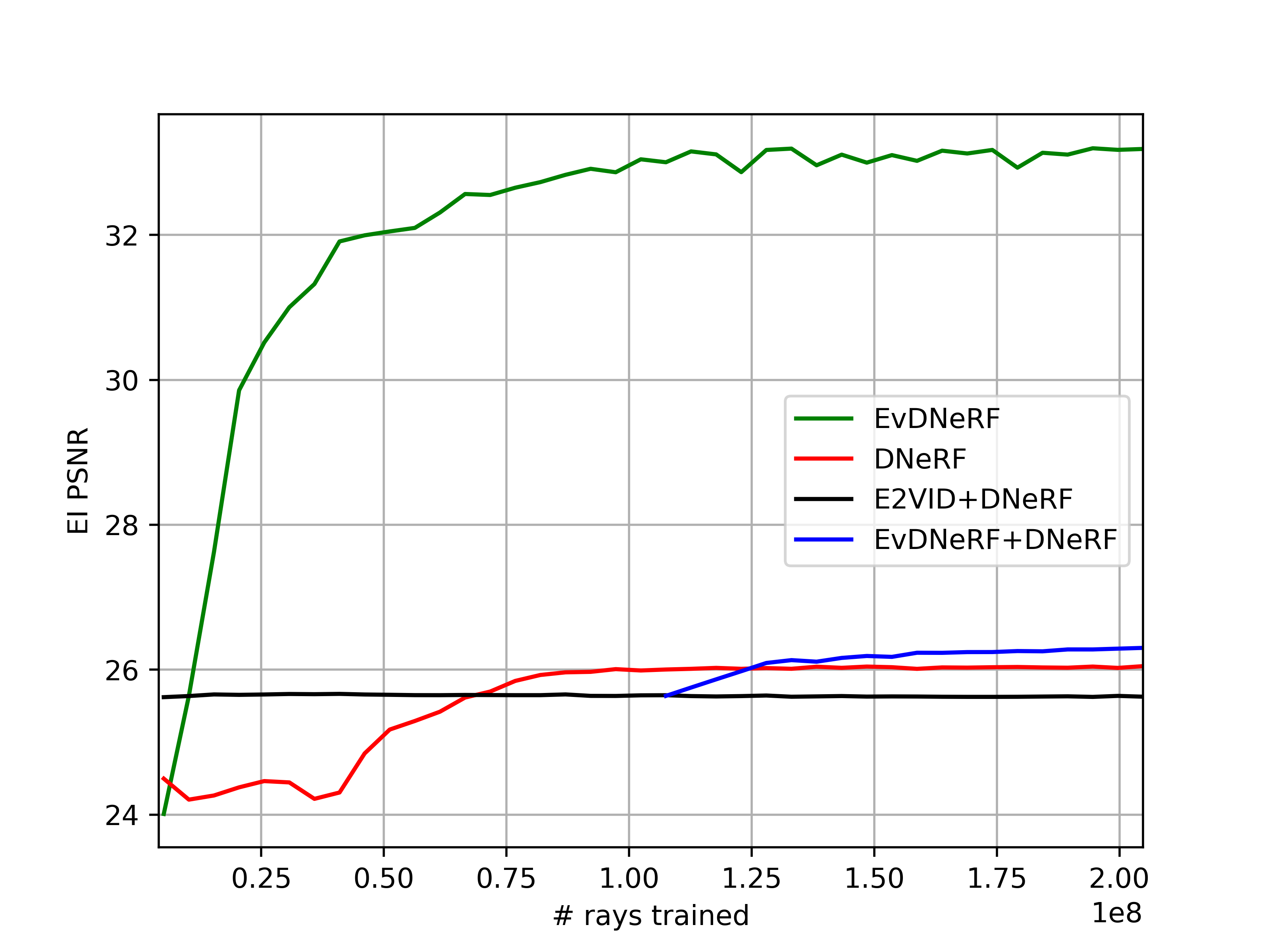}
    \caption{PSNR of event image (EI) reconstructions vs. number of rays trained, showing that \textit{EvDNeRF} widely outperforms baselines on the Jet-Down dataset. \textit{EvDNeRF+DNeRF} tracks \textit{EvDNeRF} exactly until the halfway point, where switching to optimizing intensity images causes a drop in EI PSNR.}
    \label{fig:ei-psnr-curve-vs-rays}
\end{figure}

\noindent \textbf{Simulated Jet scenes.} We evaluate our method compared to baselines on simulated scenes with the Jet object and three trajectories; quantitative metrics are shown in Table \ref{tab:evim-recon}, where EvDNeRF outperforms all baselines on events reconstruction PSNR (peak signal-to-noise ratio) and LPIPS (learned perceptual image patch similarity) \cite{zhang2018unreasonable}, and either outperforms or is competitive for SSIM (structural similarity) and MAE (mean absolute error) measures. Note that these metrics are calculated directly on the \mbox{$C^{\pm}$-quantized} values of $\Delta \hat{L}$ when possible; for example, the data range for PSNR is calculated as the range of ground truth $\Delta L_{t_k}$ calculated over all pixels from Equation \ref{eq:events-to-delta-L}. Figure \ref{fig:ei-psnr-curve-vs-rays} demonstrates that EvDNeRF quickly approaches very high PSNR (34) for event reconstruction during training, while baseline methods cannot surpass a ceiling of 27.

\begin{figure}
    \centering
    \includegraphics[width=1.0\linewidth]{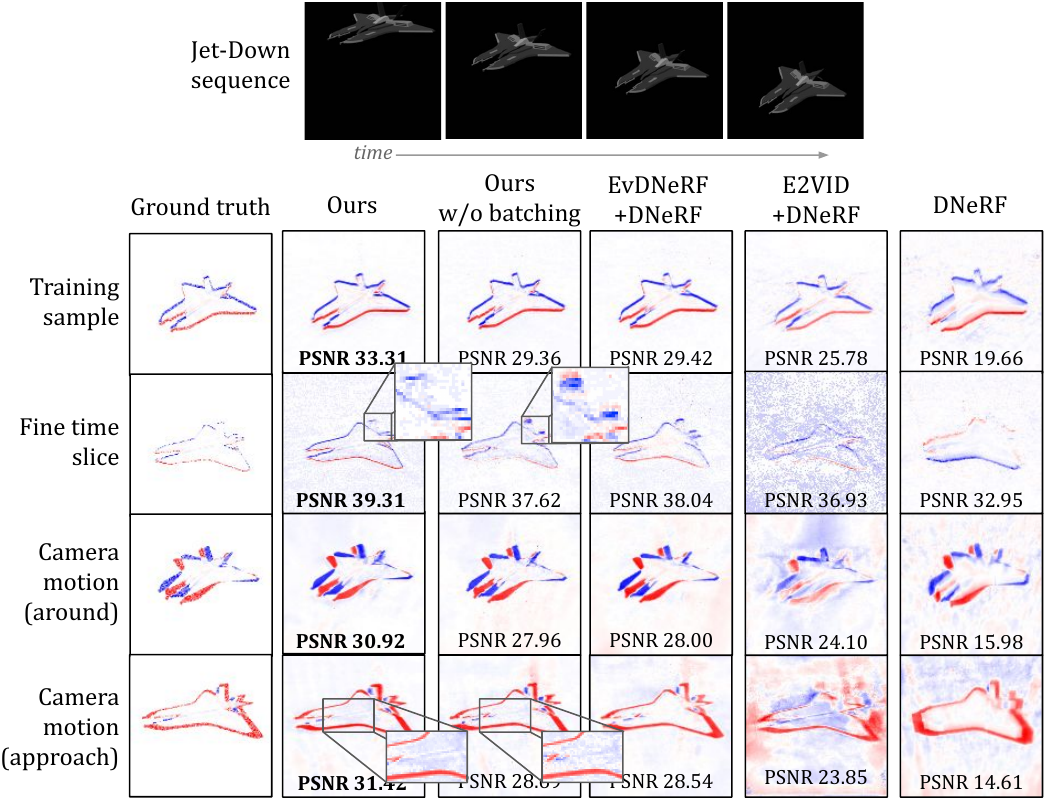}
    \caption{Comparisons of predicted event images between our method, an ablation without our varied events batching, and baselines, for various test-time queries including unseen temporal resolutions and camera motions. Our method constructs events with the best temporal resolution, as seen in the images as sharp object ``edges'' and fine details. Our reconstructions also do not contain as much spurious background predictions as DNeRF-based methods. Note that since we compare event frames with the un-normalized $\Delta \hat{L}$ values, the PSNR values should only be considered a relative metric, rather than an absolute metric as is commonly used for uint8-valued images.}
    \label{fig:evim-analysis}
\end{figure}

Variations of unseen time windows and camera motions, as well as a training sample for comparison, can be seen in Figure \ref{fig:evim-analysis}. In addition to the baselines mentioned above, we include an ablation of our method without varied batching of events during training. Across all samples, EvDNeRF with varied event batching (\textit{Ours}) has the highest PSNR on event images. The first notable improvements EvDNeRF makes over DNeRF methods is apparent in the training sample reconstruction (first row), where EvDNeRF both (a) has a sharper temporal distribution of events (represented by a sharper object), and (b) has fewer background event predictions due to the positive sampling methodology of events rather than randomly sampling rays in the image. When testing on a very fine time window (second row), the ground truth sample appears patchy along the edges of the jet; this is an artifact of the VID2E procedure for generating events. EvDNeRF, however, provides a smooth and consistent eventstream in the same areas, further demonstrating an edge of our method over existing simulators. Direct comparison to the non-batching ablation also shows that our method better reconstructs fine details of the jet fins (shown in cutouts).

Though training data only contains stationary camera viewpoints, the bottom two rows query events during camera motion. In these examples, we again see that \textit{E2VID+DNeRF} and \textit{DNeRF}, which both only use DNeRF to train, have more severe, spurious background predictions. While our method boasts the highest PSNR values, visually some of the methods' reconstructions appear similar. However, the bottom cutouts (with slightly improved contrast) show that our method reconstructs fine events on the interior of the jet--the two faint lines running along the length of the jet. These are less clear in the other methods. More variety of camera viewpoints are shown in the supplementary.

Note that since \textit{EvDNeRF+DNeRF} outperforms \textit{Ours w/o batching} in terms of PSNR, it might suggest we should apply batching to \textit{EvDNeRF+DNeRF} for fair comparison to EvDNeRF. However, temporal batching is not applicable to the highly discretized images, which arrive at a fixed frequency (32Hz); batching can only be done with events since the datastream can be considered near-continuous (MHz+).


\begin{table*}[t]
\resizebox{\linewidth}{!}{%
\centering
\begin{tabular}{rccclcccclcccclcccclc}
\multicolumn{1}{l}{}          & \multicolumn{20}{c}{{\ul Simulated data event image reconstruction}}                                                                                                                                                                                                                                                                                                                                                         \\
\multicolumn{1}{l}{\textbf{}} & \multicolumn{5}{c|}{Jet-Down}                                                                              & \multicolumn{5}{c|}{Jet-Spiral}                                                                            & \multicolumn{5}{c|}{Jet-Land}                                                                              & \multicolumn{5}{c}{Multi}                                                             \\
\multicolumn{1}{l}{}          & PSNR↑          & SSIM↑          & \multicolumn{2}{c}{LPIPS↓}         & \multicolumn{1}{c|}{MAE↓}           & PSNR↑          & SSIM↑          & \multicolumn{2}{c}{LPIPS↓}         & \multicolumn{1}{c|}{MAE↓}           & PSNR↑          & SSIM↑          & \multicolumn{2}{c}{LPIPS↓}         & \multicolumn{1}{c|}{MAE↓}           & PSNR↑          & SSIM↑          & \multicolumn{2}{c}{LPIPS↓}         & MAE↓           \\ \cline{2-21} 
DNeRF                         & 26.11          & 0.800          & \multicolumn{2}{c}{0.325}          & \multicolumn{1}{c|}{0.568}          & 25.83          & 0.786          & \multicolumn{2}{c}{0.527}          & \multicolumn{1}{c|}{0.620}          & 30.13          & 0.868          & \multicolumn{2}{c}{0.329}          & \multicolumn{1}{c|}{0.329}          & 25.57          & 0.499          & \multicolumn{2}{c}{0.662}          & 1.012          \\
E2VID+DNeRF                   & 26.97          & 0.820          & \multicolumn{2}{c}{0.237}            & \multicolumn{1}{c|}{0.448}          & 26.82          & 0.848          & \multicolumn{2}{c}{0.354}            & \multicolumn{1}{c|}{0.446}          & 30.82          & \textbf{0.871}          & \multicolumn{2}{c}{0.223}            & \multicolumn{1}{c|}{\textbf{0.258}}          & 26.83          & 0.821          & \multicolumn{2}{c}{0.377}          & 0.427          \\
EvDNeRF+DNeRF                 & 32.54          & 0.877          & \multicolumn{2}{c}{0.140}          & \multicolumn{1}{c|}{0.364}          & 31.08          & \textbf{0.878}          & \multicolumn{2}{c}{0.153}          & \multicolumn{1}{c|}{\textbf{0.384}}          & 33.08          & 0.851          & \multicolumn{2}{c}{0.176}          & \multicolumn{1}{c|}{0.338}          & 27.45          & 0.841          & \multicolumn{2}{c}{0.154}          & 0.374          \\ \cline{2-21} 
EvDNeRF (ours)                & \textbf{33.39} & \textbf{0.891} & \multicolumn{2}{c}{\textbf{0.117}} & \multicolumn{1}{c|}{\textbf{0.336}} & \textbf{32.57} & 0.848 & \multicolumn{2}{c}{\textbf{0.145}} & \multicolumn{1}{c|}{0.392} & \textbf{34.56} & 0.841 & \multicolumn{2}{c}{\textbf{0.149}} & \multicolumn{1}{c|}{0.332} & \textbf{27.50} & \textbf{0.843} & \multicolumn{2}{c}{\textbf{0.079}} & \textbf{0.366}
\end{tabular}}
\caption{\label{tab:evim-recon}Quantitative metrics comparing EvDNeRF to baselines for a simulated scene of the Jet object on three different trajectories, as well as the Multi multiple object dataset. EvDNeRF outperforms the baselines for most datasets across various metrics.}
\end{table*}

\begin{figure}
    \centering
    \includegraphics[width=0.9\linewidth]{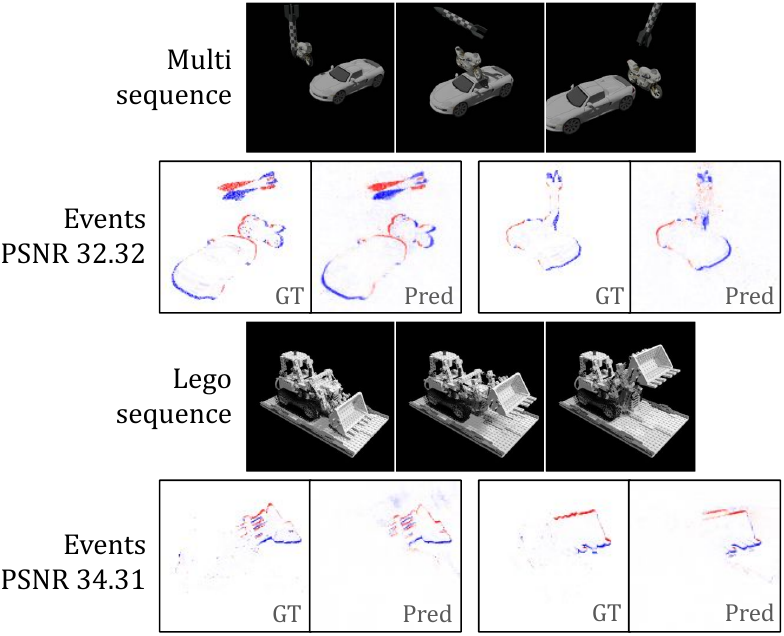}
    \caption{Reconstructions of events by EvDNeRF on simulated scenes with non-rigid deformations, at test viewpoints. Predictions (\textit{Pred}) are very similar to ground truth (\textit{GT}) and return high PSNR.}
    \label{fig:nonrigid}
\end{figure}

\noindent \textbf{Non-rigid scene deformations.} We present events reconstruction with EvDNeRF on two simulated scenes with non-rigid deformations, or scenes where viewpoint change cannot emulate scene dynamics (Figure \ref{fig:nonrigid}). Performance is comparable to the rigid deformation case of the Jet datasets. The Multi dataset features lots of occlusions and shadows; occlusions can challenge the deformation network $\Psi_t$ of the DNeRF backbone during training, but we found it to work well. The Lego tractor scene has a large stationary cabin, with only the bucket being raised mechanically by the lego components; this is why only the bucket is seen in the events reconstruction, and image and depth reconstructions of the scene also primarily show the bucket (see supplementary).

\noindent \textbf{Real world.} For demonstration on real-world data, we use a fork rotating on a servo as described in Section \ref{sec:event-data-generation}; this simple object was chosen due to its broad structure and fine features. The Real-Fork dataset eventstream contains background and residual events that are characteristic of event camera ``noise''. We believe this causes many low-valued, dispersed background events, or ``floating'' events, in the reconstructions. These are very easily filtered out with a low threshold, which we do for the event images we present in this paper in Figure \ref{fig:real-world-evims} (non-filtered images can be found in the supplementary). Reconstructions in Figure \ref{fig:fork-evims} are very good even on challenging test viewpoints such as the nearly head-on view of the fork approaching the camera (first \textit{Test views} image).

\begin{figure}
  \centering
  \begin{subfigure}{0.9\linewidth}
    \includegraphics[width=1.0\linewidth]{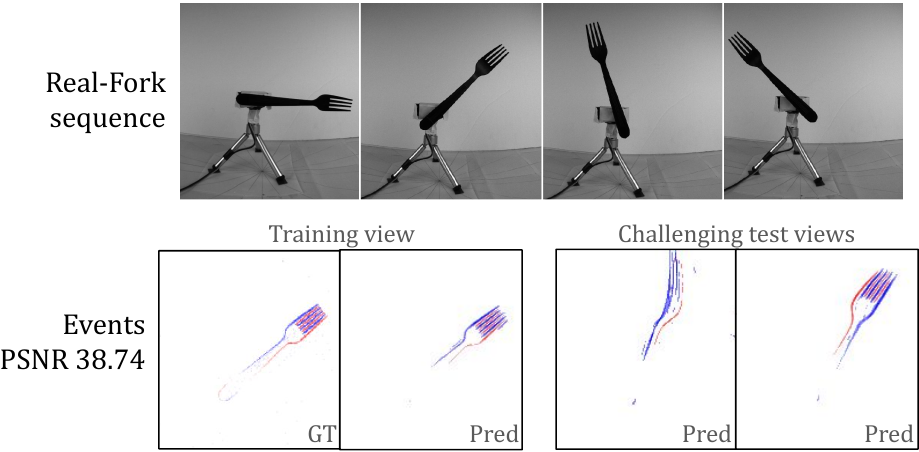}
    \caption{Images of scene motion and events reconstructions.}
    \label{fig:fork-evims}
  \end{subfigure}\vspace{.5cm}
  \begin{subfigure}{0.9\linewidth}
    \includegraphics[width=0.95\linewidth]{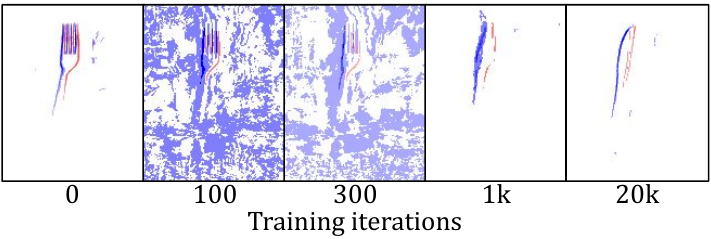}
    \caption{Visualizing the training transfer from a fork to a knife.}
    \label{fig:fork-to-knife}
  \end{subfigure}
  \vspace{0.1cm}
  \caption{Various results on Real-Fork. (a) shows a single training sample for comparison to ground truth and PSNR, as well as two challenging test samples. See the supplementary for full trajectory reconstructions. (b) shows the training progression of transferring an EvDNeRF model to a new object.}
  \label{fig:real-world-evims}
\end{figure}

\subsection{Transferring to a new object}

NeRF models are generally considered non-transferable to novel objects or scenes since they are self-supervised to learn the geometry of the scene on which it was trained. However, we explore the possibility of transferring an EvDNeRF trained on real-world data of one object to another with the same trajectory. The objects here are of roughly the same size, simplifying the transfer learning process. Figure \ref{fig:fork-to-knife} demonstrates transfer progress by showing a validation viewpoint at a few snapshots in training. Note that while training on the fork from scratch took 200k iterations, the transfer to the knife took less than 20k iterations. Since training EvDNeRF from events is more unstable than training a dynamic NeRF from images, such a transfer might be very useful to reduce train time on new, but similar, scenes.

\subsection{Limitations and extensions}

\begin{figure}
    \centering
    \includegraphics[width=1.0\linewidth]{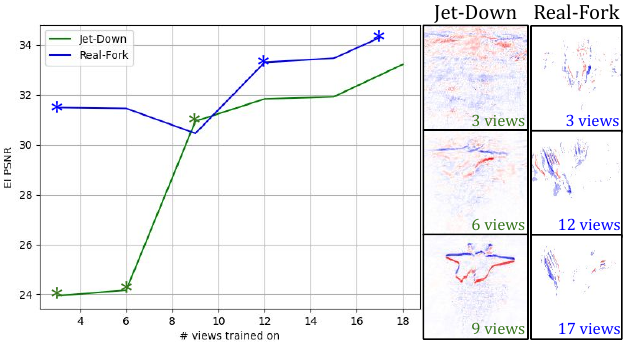}
    \caption{Events reconstruction PSNR as a function of the number of views trained on, for a simulated and real dataset after partial training of EvDNeRF. Validation reconstructions are extracted at \textit{*} points and shown on the right. For Jet-Down, 9 views are necessary to correctly place the object's events reconstructions in space. For Real-Fork, more views appears to better constrain reconstructions in the temporal dimension.}
    \label{fig:psnr-v-numviews}
\end{figure}

We find that number of training views has a big impact on events reconstruction to constrain both the spatial and temporal dimensions of dynamic scenes. Figure \ref{fig:psnr-v-numviews} shows the events reconstruction PSNR at an intermediate point in training as number of training views increases. We show this for the simulated Jet-Down dataset and the Real-Fork dataset. As seen in the Jet-Down reconstructions on the right of the figure, captured from the * points on the green curve, 9 training views correctly positions the object in the center of the frame, whereas fewer views does not constrain the position of the object in space. For Real-Fork, the object has reduced spread across its trajectory as number of views increases, indicating improved inference of the temporal dynamics of the scene. These results show that many views are needed for training an accurate EvDNeRF, as enabled by our simulated and real data generation pipelines.


While this work focuses on developing an eventstream simulator that can generalize to novel views and temporal resolutions using a NeRF backbone, the intensity image and depth map reconstruction from EvDNeRF can be poor. Intensity images can be affine-shifted in log brightness to match that of ground truth images, but the visual quality of these images is not as good as those seen from event-trained static-NeRF works (see Section \ref{sec:related-works}). This may be because the DNeRF backbone attempts to separate the temporal and spatial dimensions via two MLPs and only uses reconstruction loss terms, a difficult task made more challenging by training directly from events. It is possible that utilizing additional consistency loss terms to supervise flow \cite{du2021neural, li2021nsff} or combining with flow estimation from events \cite{gehrig2021raft, gehrig2022dense, zhu2018ev} might improve our results. We may also use a different backbone that jointly learns the time-spatial relationship within one architecture \cite{park2021hypernerf, li2021nsff}. Further work will be necessary to determine whether these improve intensity image and depth reconstruction when training directly from event data.

\section{Conclusion}
\label{sec:conclusion}






We present EvDNeRF, the first work addressing eventstream reconstruction across novel temporal resolutions and viewpoints of a dynamic scene trained solely from events, by leveraging the strong multi-view consistency properties of neural radiance fields. We thoroughly assess the performance of our method, compared to baselines designed to address possible limitations of EvDNeRF, across various scenes including rigid and non-rigid deformations, in simulation and in real-world. One limitation is the instability of training from events, likely due to the fact that events capture differences in scene appearance rather than the scene itself; however, our experiment on transferring EvDNeRF models to new scenes suggests that training on new, similar scenes can be done quickly, reducing training time and likelihood of divergence.

Since EvDNeRF can be used as an events simulator of a given scene, where novel event batches unseen in the given dataset can be queried, this work might enable future robotics and computer vision tasks or research where in-situ event data may not be easily acquired. Future work leveraging NeRFs for simulating eventstreams may apply more recent advances in dynamic NeRFs to events, such as unbounded scenes and sparse-view training; jointly utilizing blurred image data with eventstreams may improve image and depth reconstruction; other work may learn the controllable dynamics or object properties in a dynamic scene.


\section*{Acknowledgements}
\label{sec:acknowledgements}

We would like to thank all the members of the Autonomous Systems Research group at Microsoft Research for their support and discussions; Anthony Bisulco for guidance on collecting real eventstream data; Bernd Pfrommer for his excellent work on drivers and software supporting event camera research; Jiahui Lei for NeRF-related suggestions and discussions; Kenneth Chaney for his support on using event cameras and related software. This work was supported by National Science Foundation, grant no. DGE-2236662.


{\small
\bibliographystyle{ieee_fullname}
\bibliography{egbib}
}

\end{document}


\title{Supplementary\\EvDNeRF: Reconstructing Event Data with Dynamic Neural Radiance Fields}
\maketitle

\section{Code, datasets, and pretrained weights}

Code, datasets, and pretrained weights can be found here: \url{https://github.com/anish-bhattacharya/EvDNeRF}.

\begin{figure*}
    \centering
    \includegraphics[width=0.8\linewidth]{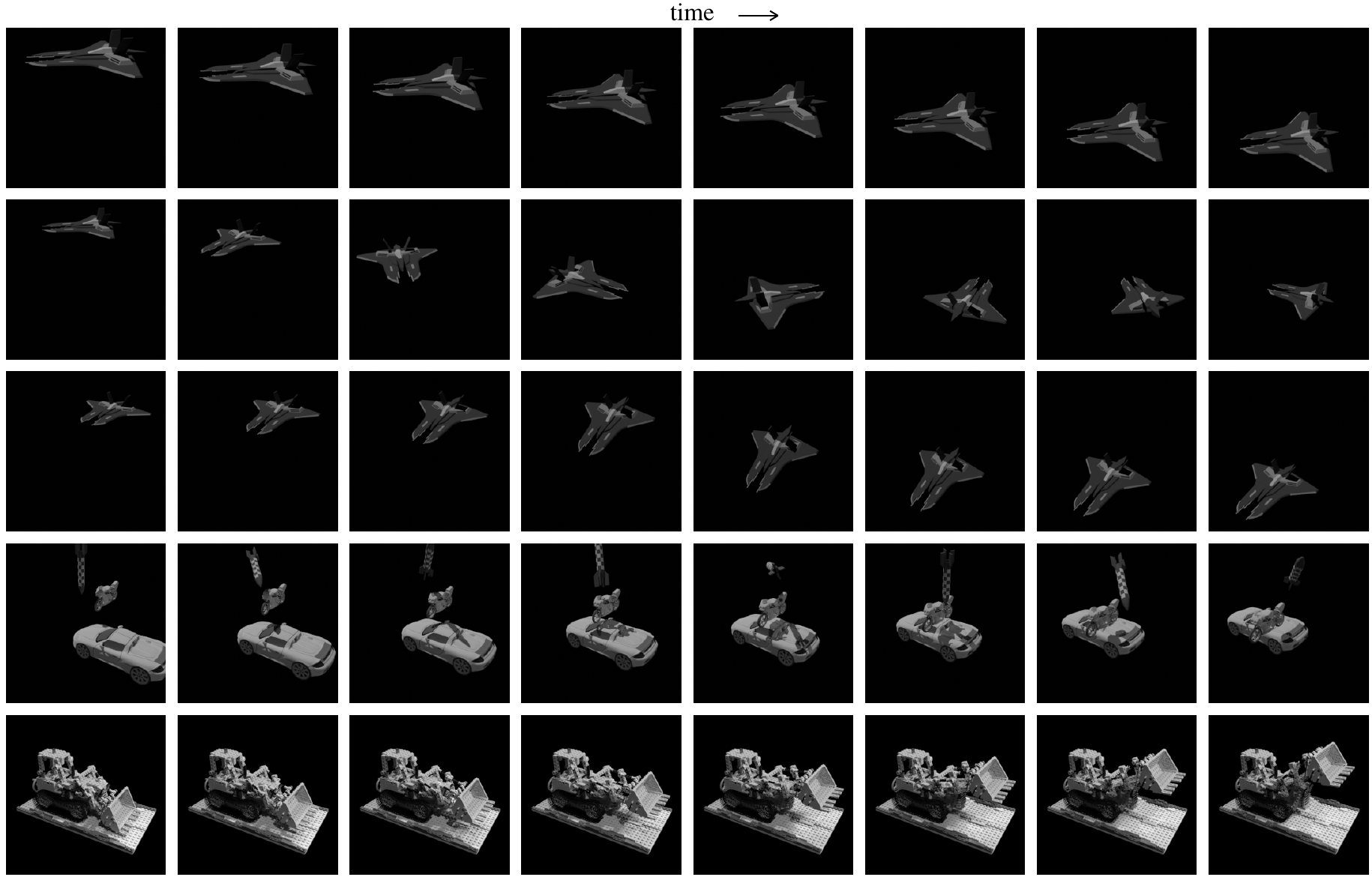}
    \caption{Ground truth intensity image snapshots of all simulated datasets: Jet-Down, Jet-Spiral, Jet-Land, Multi, and Lego; time progresses from left to right.}
    \label{fig:sim-datasets-ims}
\end{figure*}

\section{Image reconstruction}

\begin{figure}
    \centering
    \includegraphics[width=0.9\linewidth]{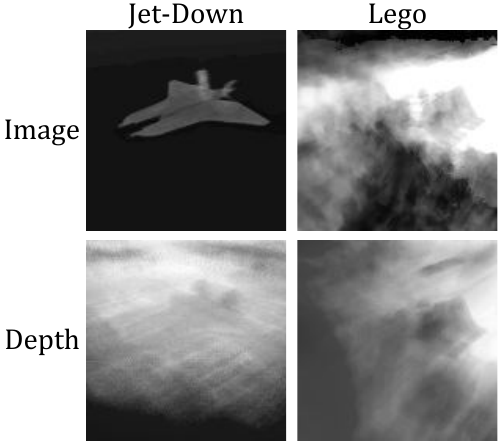}
    \caption{Sample image and depth reconstructions of Jet-Down and Lego datasets, at test viewpoints. Quality of these reconstructions is low, as we expect since we design EvDNeRF to only predict events well. But the objects can be made out, and it is possible that some tuning may improve results. Images were shifted to match ground truth brightness, and depth maps were contrast adjusted.}
    \label{fig:imrecons}
\end{figure}

\begin{figure}
    \centering
    \includegraphics[width=0.9\linewidth]{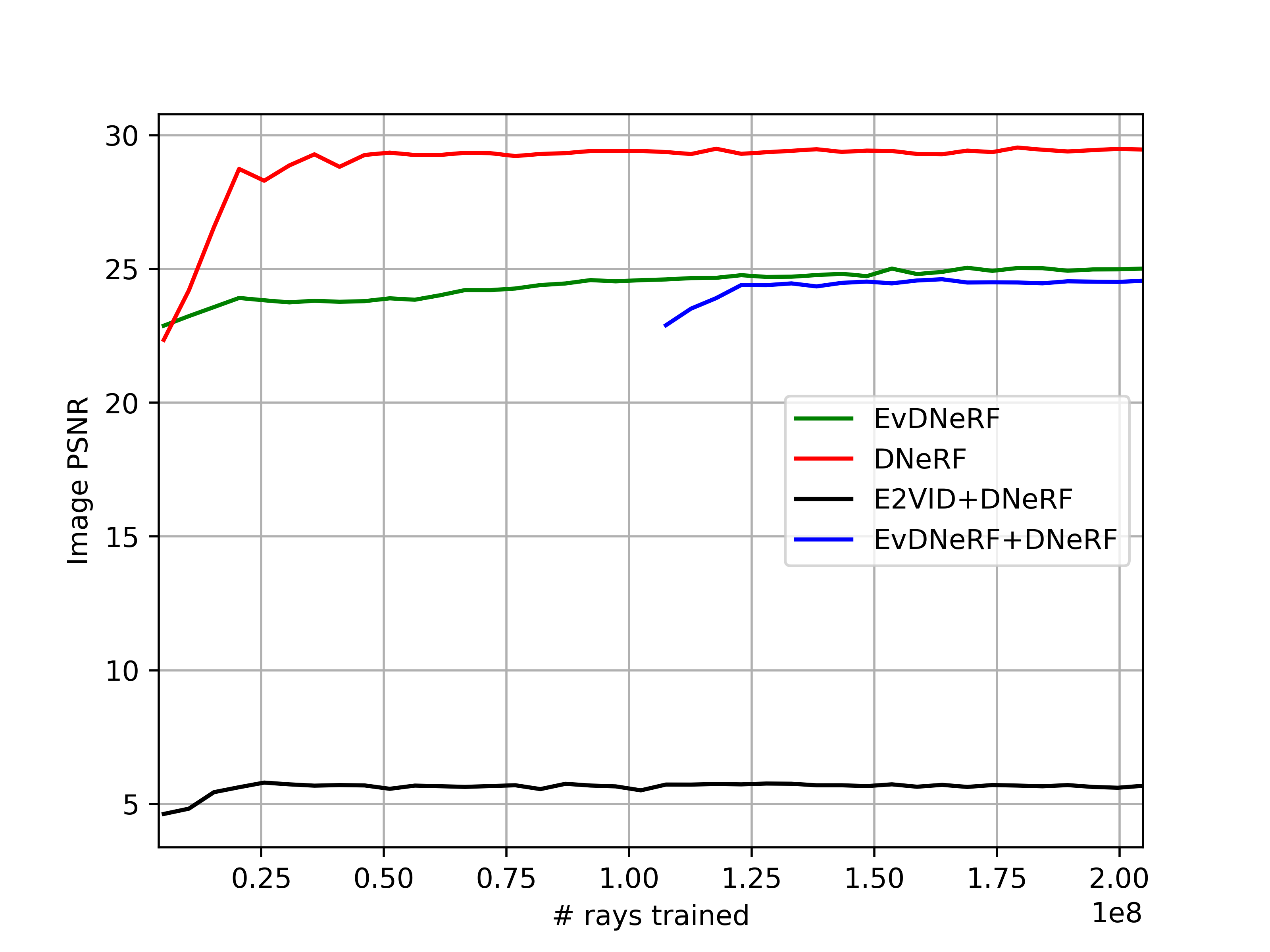}
    \caption{PSNR curves of intensity image generation of our method versus baseline methods, on the Jet-Down dataset (note that these curves are without brightness shifting, whereas tabulated metrics are with brightness adjustment).}
    \label{fig:image-psnr-curve-vs-rays}
\end{figure}

We design EvDNeRF to predict events reconstructions well, and do not expect it to perform well on absolute intensity images or depth of the scene. Training purely from events data does not provide complete scene information to achieve good absolute-value reconstructions. However, for completeness and discussion, we present examples of image and depth reconstructions by EvDNeRF in Figure \ref{fig:imrecons}, and quantitative metrics for image reconstruction in Table \ref{tab:im-recon} and Figure \ref{fig:image-psnr-curve-vs-rays}. The objects can be made out by the human eye, but there are floating artifacts, loss of fine details, and background depth inaccuracies. The problem of image reconstruction is challenging since we are only training from events; this is especially noticeable in the loss of fine details in the image of the jet and in the poor quality of the stationary cabin of the lego tractor (the only supervision for training EvDNeRF on the cabin is when the bucket passes in front of the cabin, triggering events dependent on the bucket-cabin contrast). A previous work (Klenk, et al. (2023)) used RGB data jointly with events to constrain the color predictions, which might help here. Additional loss terms constraining flow and regularizing density values throughout the scene might improve these aspects of EvDNeRF performance.

\begin{table}[]
\resizebox{\linewidth}{!}{%
\begin{tabular}{rcccc}
\multicolumn{1}{l}{\textbf{}} & \multicolumn{4}{c}{{\ul Jet-Down intensity image reconstruction}} \\
\multicolumn{1}{l}{}          & PSNR↑          & SSIM↑          & LPIPS↓          & MAE↓          \\ \cline{2-5} 
DNeRF                         & 27.43          & 0.873          & 0.183           & 0.012         \\
E2VID+DNeRF                   & 24.50          & 0.533          & 0.262           & 0.028         \\
EvDNeRF+DNeRF                 & 27.13          & 0.807          & 0.140           & 0.013         \\
EvDNeRF (ours)                & 26.98          & 0.835          & 0.138           & 0.013        
\end{tabular}}
\caption{Quantitative metrics for intensity image reconstruction, comparing EvDNeRF to baselines for the simulated Jet-Down dataset.}
\label{tab:im-recon}
\end{table}

\section{E2VID+DNeRF+VID2E}

\begin{figure}
    \centering
    \includegraphics[width=0.9\linewidth]{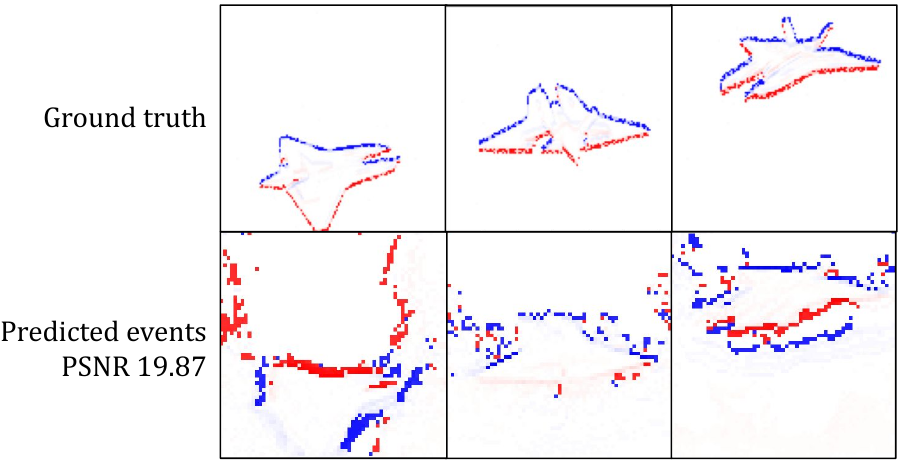}
    \caption{Event reconstructions of the \textbf{baseline} \textit{E2VID+DNeRF} but using VID2E to reconstruct events. Since these results are very poor with high levels of background events, we do not present a \textit{E2VID+DNeRF+VID2E} baseline in the paper.}
    \label{fig:vid2e-reconstruction}
\end{figure}

As mentioned in the paper, since our goal is to create an events simulator, it follows that we may explore using VID2E on top of standard DNeRF methods as a competitive baseline. However, we found that VID2E actually generated strong background events from the DNeRF predictions from \textit{E2VID+DNeRF}, as shown in Figure \ref{fig:vid2e-reconstruction}. It is possible that regularizing background events or those predicted in static areas of the scene, similar to Klenk, \etal (2023), might improve results on this particular baseline, though it would require some tuning.

\section{Implementation details}

\noindent \textbf{Implementation details.} Our code for EvDNeRF is a modified version of the DNeRF open-source code (\url{github.com/albertpumarola/D-NeRF} (2021)). As in the original NeRF implementation, two NeRF MLPs for both coarse and fine sampling are used for training EvDNeRF on real data, but results were found to be satisfactory with one NeRF model on simulated data. For paper results, we train every model for 200k iterations on a single Tesla V100 GPU (taking 1-2 days), where each iteration samples 1024 rays, 50\% of which are cast through event-triggered pixels and 50\% through random pixels. Furthermore, these rays are uniformly selected from all available viewpoints upon each training iteration, which we observed results in a more stable training than selecting all rays from a single viewpoint. Further training details follow.

We implement a learning rate warm-up for the first 1k iterations of training, train only on a cropped portion of event batches centered on high populations of events for 2k iterations, and progressively introduce scene timesteps over the first 20k iterations. This scheduling significantly improved training stability. For varied batching of our supervisory eventstream, we halve the time window of batch sizes twice, once at 100k and again at 150k of training.
 

\section{Real world data}

\subsection{Notes on time-synced real data generation with many views}

Generation of real-world data from multi-view event cameras with a consistent motion was a challenging task. While the servo-actuated motion and physical placement of the dynamic object relative to the camera was carefully calibrated, it is easily prone to small errors in position or time-synchronization. This is another key benefit of using a neural implicit representation for an events simulator; MLPs can learn smooth functions from noisy input data (within some bounds). This is also why we were able to manually time-synchronize the eventstreams using approximate calculations of motion timestamps (see Figure 3b in paper). Initial attempts to generate real-world data was done with three hardware time-synchronized event cameras, rigidly arranged to be 25$^{\circ}$ apart from each other, along a circle's arc with 30cm radius to the target. However, as noted in Section 4.3 of the paper, events reconstructions improved as number of views increased, and the front-facing three-view data was not sufficient to constrain the spatial geometry constructed by EvDNeRF. In this case, training views overfitted to achieve low training loss, but intermediate validation viewpoints returned poor reconstructions with multiple hallucinated objects in the renderings (similar to the poor positioning of the Jet in the sample events reconstructions in Figure 9 in the paper). Again, additional density or flow consistency loss terms might improve performance on number-of-views-limited, front-facing datasets.

\subsection{Filtering events reconstructions}

\begin{figure}
    \centering
    \includegraphics[width=0.95\linewidth]{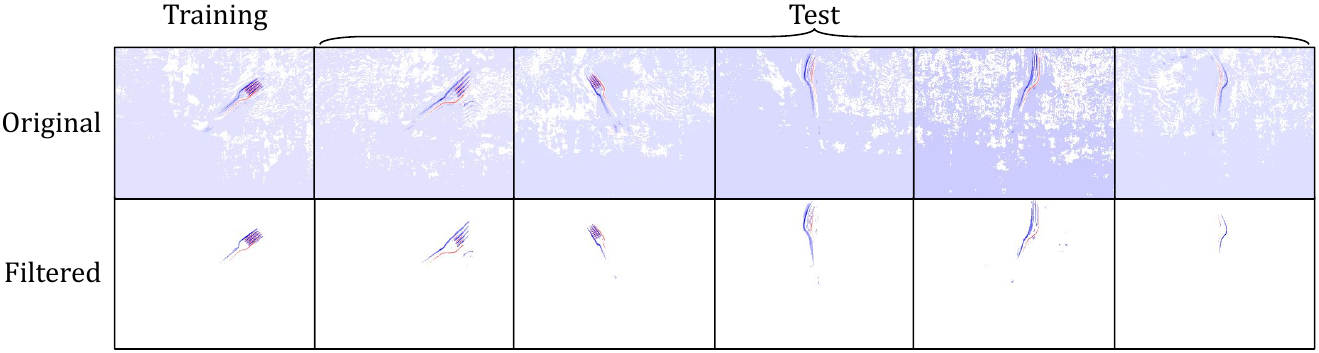}
    \caption{Comparison of original events reconstructions on Real-Fork by EvDNeRF and filtered versions.}
    \label{fig:filtering}
\end{figure}

As mentioned in Results, we filter out low-valued events from real-world reconstructions. Comparisons between original and filtered events are shown in Figure \ref{fig:filtering}. Note that the true, object-triggered events are consistently high valued, and therefore are easy to filter from the background. It's possible that the low levels of background event noise in the Real-Fork dataset cause the consistent levels of background event predictions; however, simple attempts to filter out event noise in the gathered data (via a median filter) caused divergence of EvDNeRF training, likely since we violate brightness consistency of underlying image predictions by manipulating the events in such a way.

\section{Additional event reconstruction results}

\begin{figure*}
    \centering
    \includegraphics[width=1.0\linewidth]{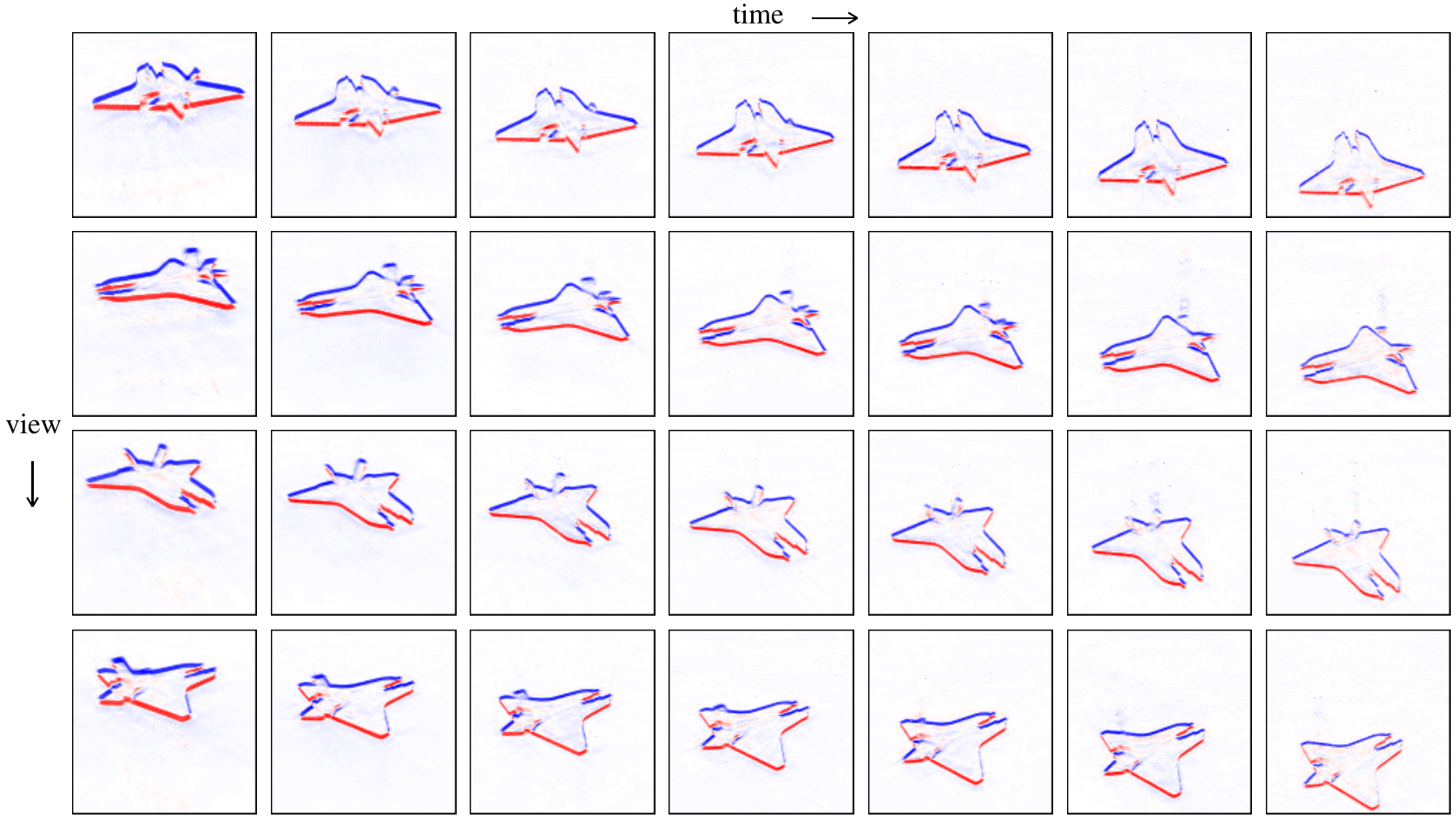}
    \caption{Jet-Down dataset: event reconstructions across time and viewpoint.}
    \label{fig:jd-evims}
\end{figure*}

\begin{figure*}
    \centering
    \includegraphics[width=1.0\linewidth]{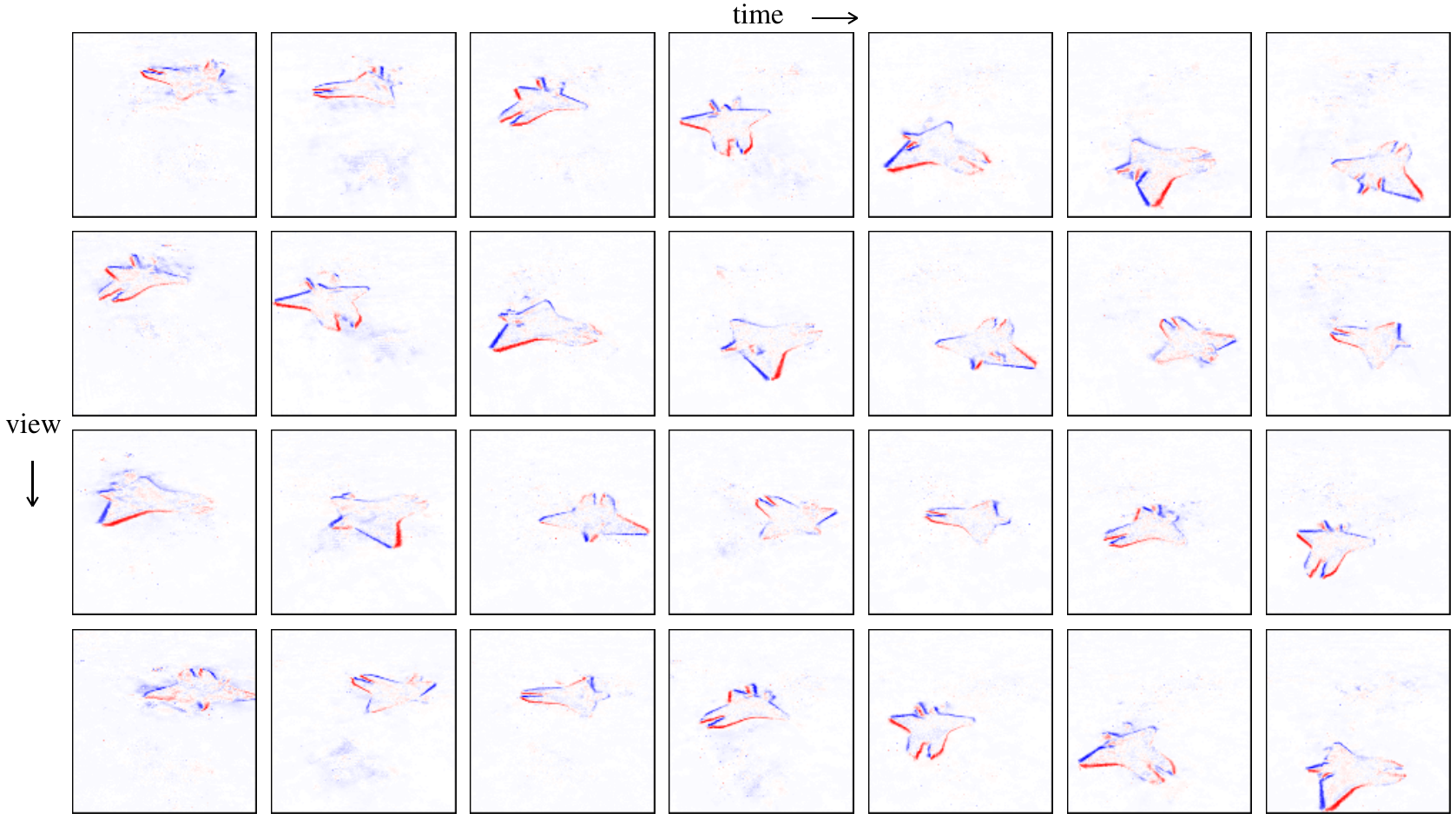}
    \caption{Jet-Spiral dataset: event reconstructions across time and viewpoint.}
    \label{fig:js-evims}
\end{figure*}

\begin{figure*}
    \centering
    \includegraphics[width=1.0\linewidth]{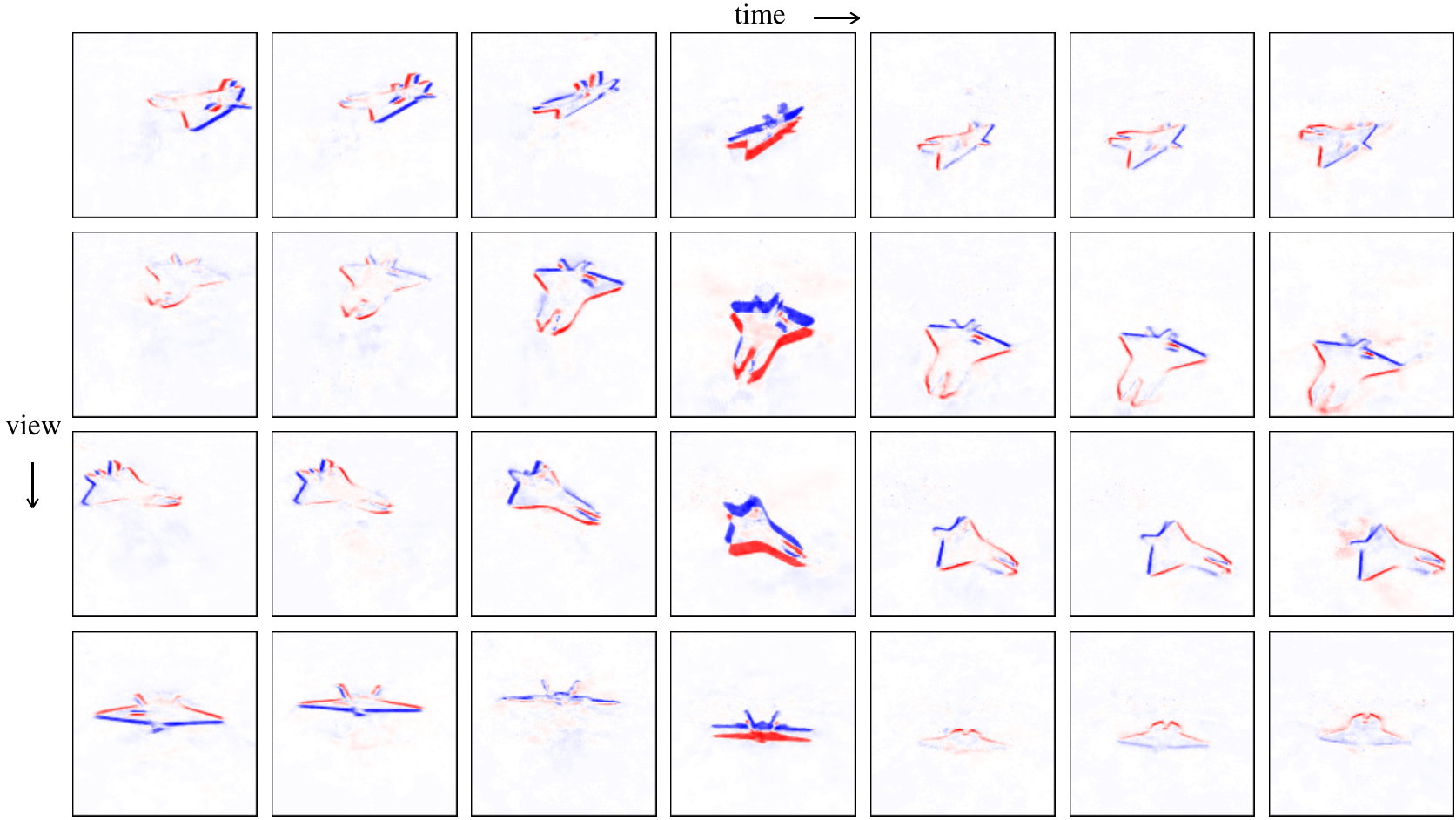}
    \caption{Jet-Land dataset: event reconstructions across time and viewpoint.}
    \label{fig:jl-evims}
\end{figure*}

\begin{figure*}
    \centering
    \includegraphics[width=1.0\linewidth]{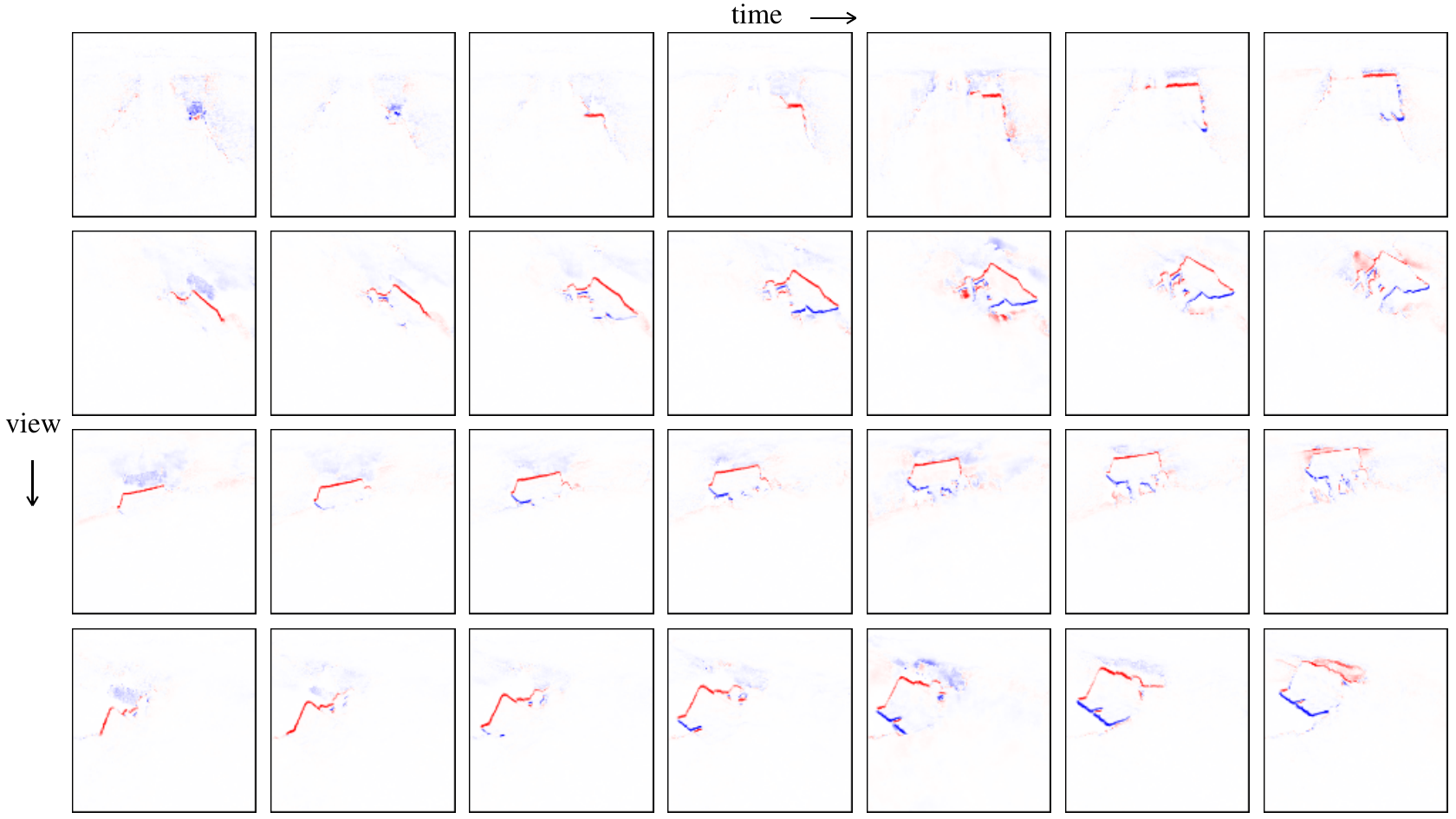}
    \caption{Lego dataset: event reconstructions across time and viewpoint.}
    \label{fig:lego-evims}
\end{figure*}

\begin{figure*}
    \centering
    \includegraphics[width=1.0\linewidth]{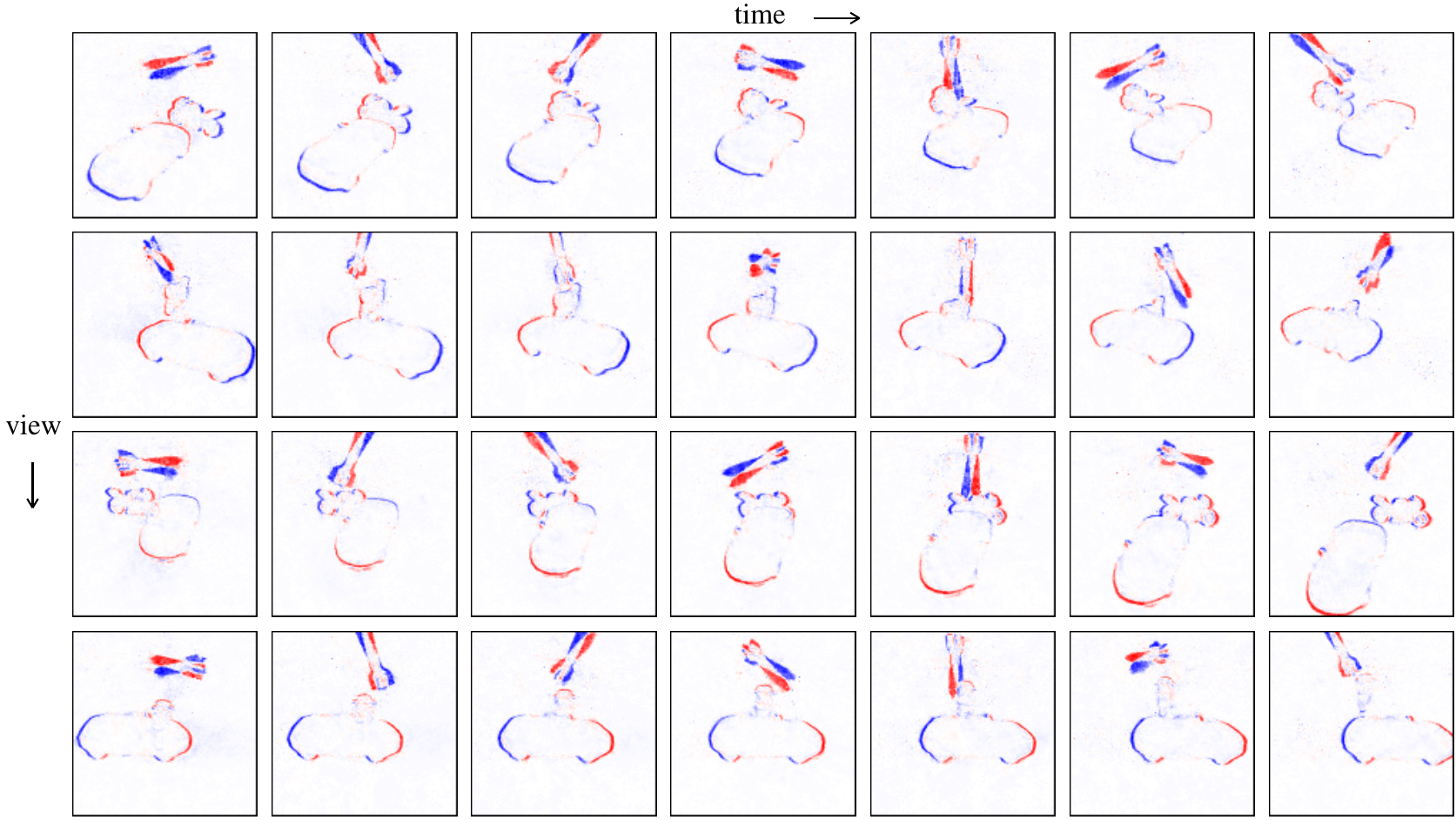}
    \caption{Multi dataset: event reconstructions across time and viewpoint.}
    \label{fig:multi-evims}
\end{figure*}

\begin{figure*}
    \centering
    \includegraphics[width=1.0\linewidth]{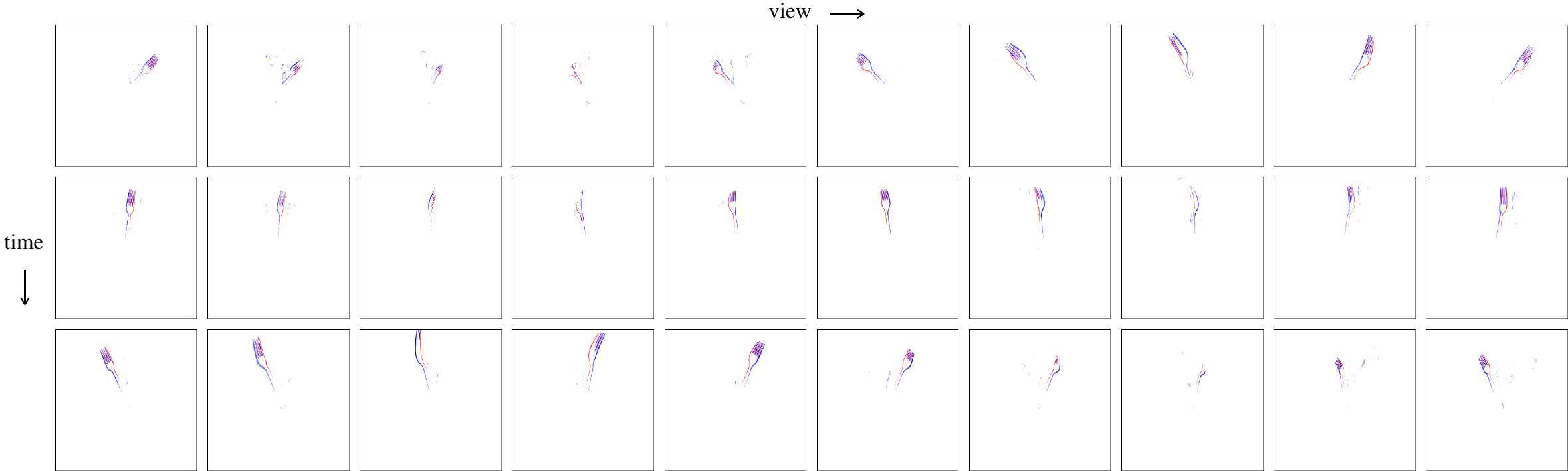}
    \caption{Real-Fork dataset: event reconstructions across time and viewpoint. Note that view and time axes are swapped for this figure.}
    \label{fig:real-fork-evims}
\end{figure*}

We present additional event reconstructions across test time windows and viewpoints. See Figures \ref{fig:jd-evims} - \ref{fig:real-fork-evims}.

